%% file: example_paper.tex
\documentclass[nohyperref]{article}

\usepackage{microtype}
\usepackage{graphicx}
\usepackage{subfigure}
\usepackage{booktabs} 

\usepackage{hyperref}

\usepackage[accepted]{icml2022}

\usepackage{amsmath}
\usepackage{amssymb}
\usepackage{mathtools}
\usepackage{amsthm}

\usepackage[capitalize,noabbrev]{cleveref}

\theoremstyle{plain}

\theoremstyle{definition}

\theoremstyle{remark}

\usepackage[textsize=tiny]{todonotes}

\usepackage{amsmath}
\usepackage{amsthm}
\usepackage{amssymb}

\usepackage{color}
\usepackage{xcolor}
\usepackage{subfigure}
\usepackage{bm}

\usepackage{colortbl}

\definecolor{mygray}{gray}{.9}

\definecolor{mypink}{rgb}{.99,.91,.95}

\definecolor{mycyan}{cmyk}{.3,0,0,0}

\def\ie{\textit{i.e.}}

\usepackage{algorithm}
\usepackage{algorithmic}

\graphicspath{{figs/}}
\usepackage{multirow}

\def\mb{\mathbf}
\def\mbb{\mathbb}
\def\mc{\mathcal}

\usepackage{newfloat}
\usepackage{listings}
\floatstyle{ruled}
\newfloat{listing}{tb}{lst}{}
\floatname{listing}{Listing}

\icmltitlerunning{Class-Incremental Lifelong Learning in Multi-Label Classification}

\begin{document}

\twocolumn[
\icmltitle{Class-Incremental Lifelong Learning in Multi-Label Classification}



\icmlsetsymbol{equal}{*}

\begin{icmlauthorlist}
\icmlauthor{Kaile Du}{equal,yyy}
\icmlauthor{Linyan Li}{equal,sch}
\icmlauthor{Fan Lyu}{yyy,comp}
\icmlauthor{Fuyuan Hu}{yyy}
\icmlauthor{Zhenping Xia}{yyy}
\icmlauthor{Fenglei Xu}{yyy}
\end{icmlauthorlist}

\icmlaffiliation{yyy}{Suzhou University of Science and Technology, Suzhou 215009, China.}
\icmlaffiliation{comp}{College of Intelligence and Computing, Tianjin University, Tianjin 300350, China}
\icmlaffiliation{sch}{Suzhou Institute of Trade \& Commerce, Suzhou 215009, China.}

\icmlcorrespondingauthor{Linyan Li}{lilinyan@szjm.edu.cn}

\icmlkeywords{Machine Learning, ICML}

\vskip 0.3in
]




\printAffiliationsAndNotice{\icmlEqualContribution.} 

\begin{abstract}
Existing class-incremental lifelong learning studies only the data is with single-label, which limits its adaptation to multi-label data.
This paper studies Lifelong Multi-Label (LML) classification, which builds an online class-incremental classifier in a sequential multi-label classification data stream.
Training on the data with  \textit{Partial Labels} in LML classification may result in more serious \textit{Catastrophic Forgetting} in old classes.
To solve the problem, the study proposes an Augmented Graph Convolutional Network (AGCN) with a built Augmented Correlation Matrix (ACM) across sequential partial-label tasks.
  The results of two benchmarks show that the method is effective for LML classification and reducing forgetting.
\end{abstract}

\input{intro}
\input{method}
\input{ex}

\section{Conclusion}
LML classification is a new paradigm of lifelong learning.
The key challenges are constructing label relationships and reducing catastrophic forgetting to improve overall performance.
In this paper, a novel AGCN based on an auto-updated expert mechanism is proposed to solve the challenge.
We construct a label correlation matrix with soft labels generated by an expert network. 
We also mitigate relationship forgetting by a proposed relationship-preserving loss. 
In general, AGCN connects previous and current tasks on all seen classes in LML classification.
Extensive experiments demonstrate that AGCN can capture well the label dependencies and effectively mitigate the catastrophic forgetting, thus achieving better classification performance.




{
\bibliography{example_paper}
\bibliographystyle{icml2022}}



\end{document}

%% file: intro.tex
\section{Introduction}

Class-incremental learning~\cite{rebuffi2017icarl} constructs a unified evolvable classifier, which online learns new classes from a sequential image data stream and achieves multi-class classification for the seen classes.
For privacy, storage and efficient computation reasons, the training data in lifelong learning for the old tasks are often unavailable when new tasks arrive, and the new task data only has labels of itself.
Thus, the catastrophic forgetting~\cite{kirkpatrick2017overcoming}, \ie, the training on new tasks may lead to the old knowledge overlapped by the new knowledge, is the main challenge of Lifelong Single-Label (LSL) classification. 
To solve the catastrophic forgetting problem, the state-of-art methods for class-incremental lifelong learning can be categorized into regularization-based methods, such as EWC~\cite{kirkpatrick2017overcoming}, LwF~\cite{li2017learning} and LIMIT~\cite{zhou2022few}; rehearsal-based methods, such as AGEM~\cite{chaudhry2018efficient}, ER~\cite{rolnick2019experience}, PRS~\cite{kim2020imbalanced} and SCR~\cite{mai2021supervised}. 
However, most lifelong learning studies only consider single-labelled input data, which introduces significant limitations in practical applications. 

\begin{figure}[t]
	\centering
	\includegraphics[width=1.0\linewidth]{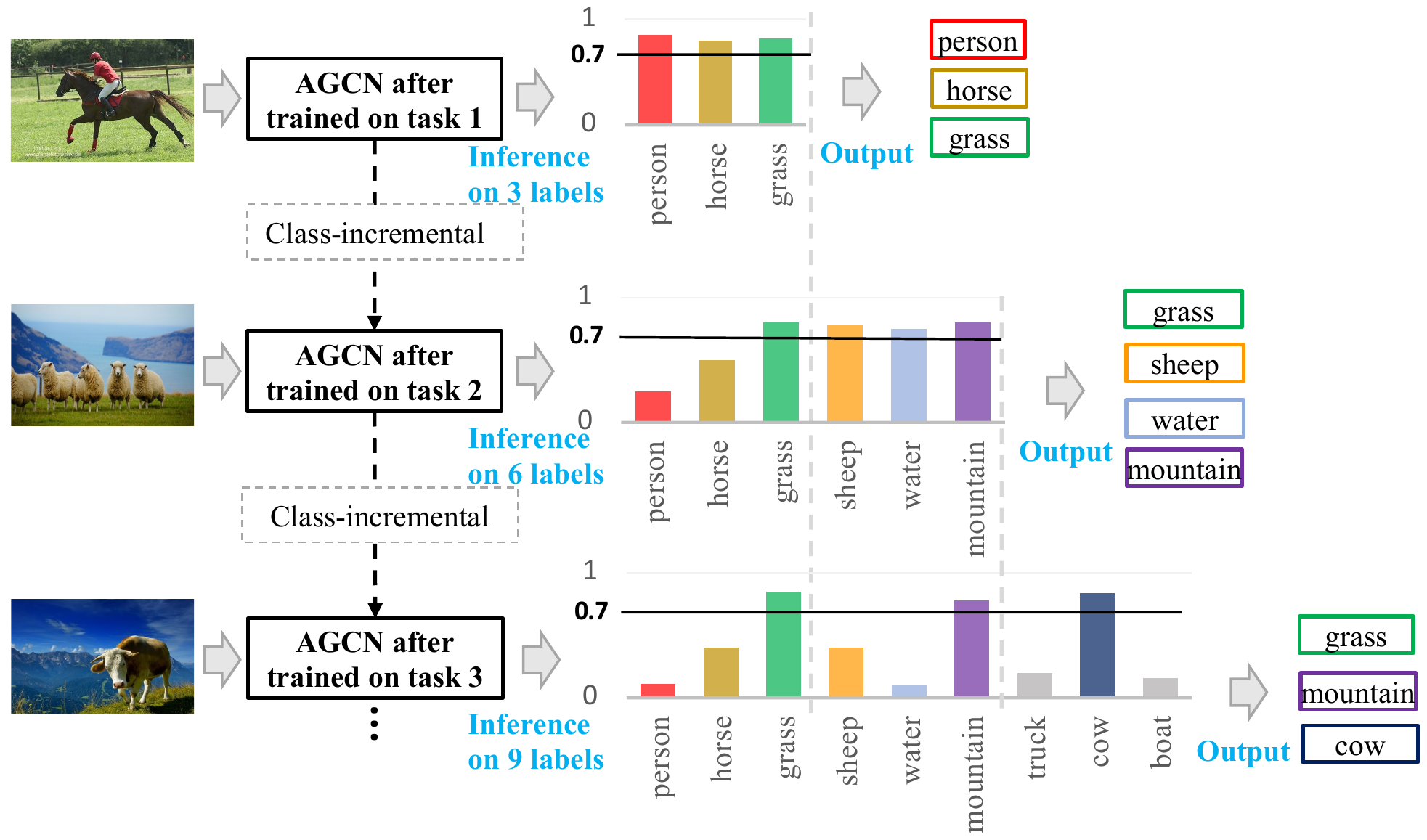}
        \vspace{-25px}
	\caption{The inference of LML classification. Given multi-label images for testing, the model can recognize more labels by learning more incremental classes (label with the probability that greater than the threshold 0.7 will be output).
	}
	\label{fig:lml}
	\vspace{-15px}
\end{figure}

This paper studies how to learn classes sequentially from new LML classification tasks. 
As shown in Fig.~\ref{fig:lml}, given testing images, and LML model can continuously recognize multiple labels with new classes learned.
It is stated that LML recognition is challenging due to not just catastrophic forgetting, but
\textit{Partial labels} for the current tasks, which means the training data may contain possible labels of past tasks.
There exist few lifelong learning algorithms designed specifically for LML against this challenge.

\begin{figure*}[t]
  \centering
  \includegraphics[width=\linewidth]{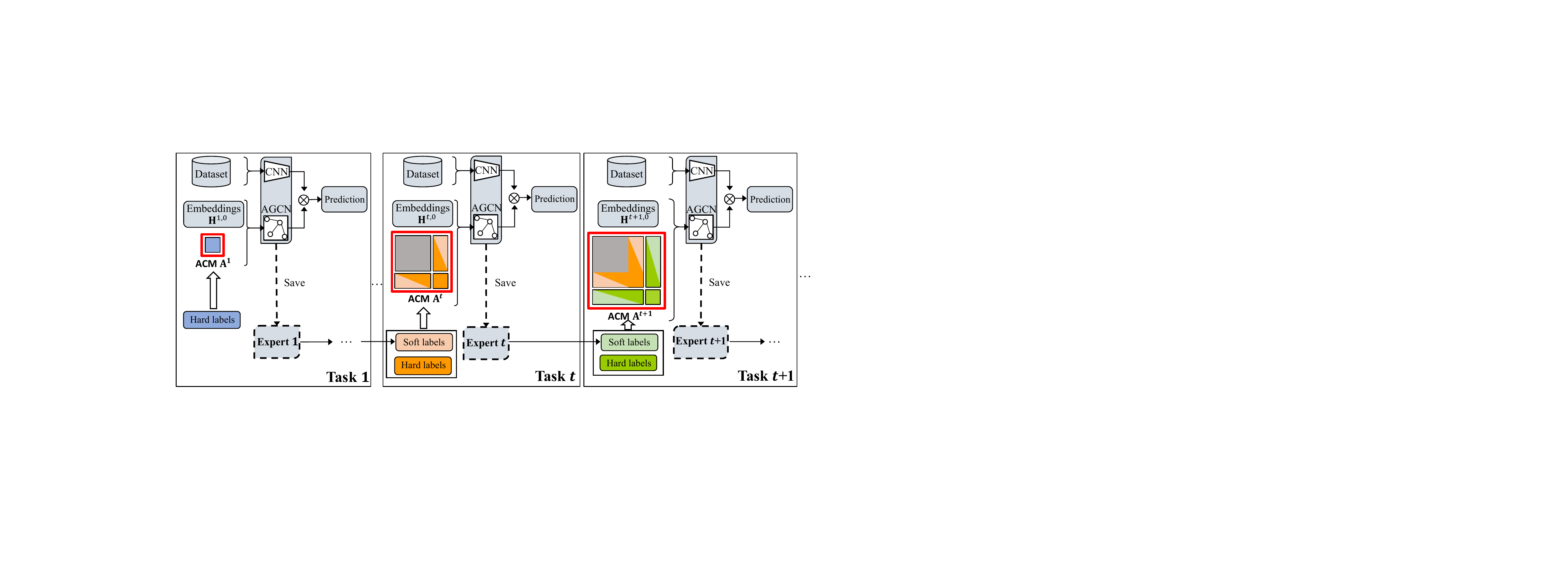}
  \vspace{-10px}
  \caption{
    The framework of AGCN. The training data for task $t$  is fed into the CNN block, and the graph node embeddings and the \textbf{ACM} are input to the AGCN block. 
    After each task has been trained, we save expert blocks to provide soft labels in the next task training. 
  }
  \label{fig:framework}
  \vspace{-15px}
\end{figure*}
We are inspired by the recent research on label relationships in multi-label learning~\cite{chen2019multi,chen2021learning}, we consider building the label relationships across tasks, \ie, label correlation matrix.
\textit{However, because of the partial label problem, it is difficult to construct the class relationships by using statistics directly.}
We propose an AGCN, a novel solution to LML classification to deal with the partial label problem. 
First, an auto-updated expert network is designed to generate predictions of the old tasks, these predictions as soft labels are used to represent the old classes for the old tasks and construct ACM.  Then, the AGCN receives the dynamic ACM and correlates the label spaces of both the old and new tasks, which continually supports the multi-label prediction. 
Moreover, to further mitigate the forgetting on both seen classes and class relationships, a distillation loss and a relationship-preserving loss function are designed for class-level forgetting and relationship-level forgetting, respectively.
We construct two multi-label image datasets, Split-COCO and Split-WIDE, based on MS-COCO and NUS-WIDE, respectively. 
The results show that our AGCN achieves state-of-art performances in LML classification.
Our code is available at \url{https://github.com/Kaile-Du/AGCN}.





%% file: method.tex
\section{Methodology}


\subsection{Lifelong multi-label learning}

In this study, each data is trained only once in the form of a data stream. 
Given $T$ recognition tasks with respect to train datasets $\{\mc{D}^{1}_\text{trn},\cdots,\mc{D}^{T}_\text{trn}\}$ and test datasets $\{\mc{D}^{1}_\text{tst},\cdots,\mc{D}^{T}_\text{tst}\}$. 
For the $t$-th task, we have new and task-specific classes to be trained, namely $\mc{C}^t$. 
\textit{The goal is to build a multi-label classifier to discriminate an increasing number of classes.}
We denote $\mc{C}_\text{seen}^t=\bigcup_{n=1}^{t}\mc{C}^n$ as seen classes at task $t$, where $\mc{C}_\text{seen}^t$ contains old class set $\mc{C}_\text{seen}^{t-1}$ and new class set $\mc{C}^t$, that is, $\mc{C}_\text{seen}^t=\mc{C}_\text{seen}^{t-1}\cup\mc{C}^t$.
Note that during the testing phase, the ground truth labels for LML classification contain all the old classes $\mc{C}_\text{seen}^t$.

\subsection{Augmented Correlation Matrix}

\label{sec:acm}

Correlation Matrix is often built for multi-label learning~\cite{chen2019multi,chen2021learning}, and can be used to construct label relationships. Our \textbf{Augmented Correlation Matrix (ACM)} provides the label relationships among all seen classes $\mc{C}^t_{\text{seen}}$ and is augmented to capture the intra- and inter-task label independences. Most existing multi-label learning algorithms~\cite{chen2019multi,chen2021learning} rely on constructing the inferring label correlation matrix $\mb{A}$ by the hard label statistics among the class set $\mc{C}$: $\mb{A}_{ij}=P(\mc{C}_i|\mc{C}_j)|_{i \neq j}$.
We construct ACM $\mb{A}^t$ for task $t>1$ in an online fashion to simulate the statistic value denoted as
\begin{equation}\label{eq:acm}
	{\mb{A}}^{t}=\begin{bmatrix} {\mb{A}}^{t-1} & \mb{R}^t \\ \mb{Q}^t & \mb{B}^t \end{bmatrix}=\begin{bmatrix} \text{Old-Old} & \text{Old-New} \\ \text{New-Old} & \text{New-New} \end{bmatrix},
\end{equation}
in which we take four block matrices including $\mb{A}^{t-1}$ and $\mb{B}^t$, $\mb{R}^t$ and $\mb{Q}^t$ to represent intra- and inter-task label relationships between old and old classes, new and new classes, old and new classes as well as new and old classes respectively.
For the first task, $\mb{A}^1=\mb{B}^1$.
For $t>1$, $\mb{A}^t\in\mbb{R}^{|\mc{C}_\text{seen}^t|\times|\mc{C}_\text{seen}^t|}$.
It is worth noting that the block $\mb{A}^{t-1}$ (Old-Old) can be derived from the old task, so we will focus on how to compute the other three blocks in the ACM.


\noindent
\textbf{New-New block} ($\mb{B}^t\in\mbb{R}^{|\mc{C}^t|\times|\mc{C}^t|}$).
This block computes the intra-task label relationships among the new classes, and the conditional probability in $\mb{B}^t$ can be calculated using the hard label statistics from the training dataset similar to the common multi-label learning: 
\begin{equation}\label{eq:hard-hard}
	\mb{B}^t_{ij} = P(\mc{C}^t_{i}\in\mc{C}^t|\mc{C}^t_{j}\in\mc{C}^t) = \frac{N_{ij}}{N_{j}},
\end{equation}
where $N_{ij}$ is the number of examples with both class $\mc{C}^t_{i}$ and $\mc{C}^t_{j}$, $N_{j}$ is the number of examples with class $\mc{C}^t_{j}$.  
Due to the online data stream, $N_{ij}$ and $N_{j}$ are accumulated and updated at each step of the training process.



\noindent
\textbf{Old-New block} ($\mb{R}^t\in\mbb{R}^{|\mc{C}_\text{seen}^{t-1}|\times|\mc{C}^t|}$). 
{Given an image $\mb{x}$, for the old classes, ${\hat{z}}_{i}$ (predicted probability) generated by the expert  can be considered as the soft label for the $i$-th class (see Eq.~\eqref{eq:distillation}). 
Thus, the product ${\hat{z}}_{i}{{{y}}_{j}}$ can be regarded as an alternative of the cooccurrences of ${\mc{C}_\text{seen}^{t-1}}_{i}$ and $\mc{C}^t_{j}$. 
$\sum_{\mb{x}}{\hat{z}}_{i}{{{y}}_{j}}$ means the online mini-batch accumulation.
Thus, we have
\begin{equation}\label{eq:soft-hard}
	\begin{aligned}		
	\mb{R}^t_{ij} 	&= P({\mc{C}_\text{seen}^{t-1}}_{i}\in{\mc{C}_\text{seen}^{t-1}}|\mc{C}^t_{j}\in\mc{C}^t) = \frac{\sum_{\mb{x}}{\hat{z}}_{i}{{{y}}_{j}}}{N_j}.
	\end{aligned}
\end{equation}
\noindent
\textbf{New-Old block} ($\mb{Q}^t\in\mbb{R}^{|\mc{C}^t|\times|\mc{C}_\text{seen}^{t-1}|}$). 
Based on  Bayes' rule, we can obtain this block by
\begin{equation}\label{eq:hard-soft}
	\begin{aligned}
		\mb{Q}^t_{ji} &= P(\mc{C}^t_{j}|{\mc{C}_\text{seen}^{t-1}}_{i}) = \frac{P({\mc{C}_\text{seen}^{t-1}}_{i}|\mc{C}^t_j)P(\mc{C}^t_j)}{P({\mc{C}_\text{seen}^{t-1}}_{i})}=\frac{\mb{R}^t_{ij}N_j}{\sum_{\mb{x}}{\hat{z}}_{i}}.
	\end{aligned}
\end{equation}
Finally, we online construct an ACM using the soft label statistics from the auto-updated expert network and the hard label statistics from the training data. 

\subsection{Augmented Graph Convolutional Network}

\label{sec:agcn}

ACM is auto-updated dependencies among all seen classes. With the established ACM, we can leverage Graph Convolutional Network (GCN) to assist the prediction of CNN as Eq.~\eqref{eq:predict}.
We propose an \textbf{Augmented Graph Convolutional Network (AGCN)} to manage the augmented fully-connected graph. AGCN is learned to map this label graph into a set of inter-dependent object classifiers. 
AGCN is a two-layer stacked graph model, which is similar to ML-GCN~\cite{chen2019multi}.
Based on the ACM $\mb{A}^t$, AGCN can capture class-incremental dependencies in an online way.
Let the graph node be initialized by the Glove embedding~\cite{pennington2014glove} namely $\mb{H}^{t,0}\in\mbb{R}^{|\mc{C}_\text{seen}^t|\times d}$ where $d$ represents the embedding dimensionality.
The graph presentation $\mb{H}^t \in\mbb{R}^{|\mc{C}_\text{seen}^t| \times D}$ in task $t$ is mapped by:
\begin{equation}\label{eq:agcn}
		\mb{H}^t=\text{AGCN}(\mb{A}^t, \mb{H}^{t,0}).
\end{equation}

As shown in Fig.~\ref{fig:framework}, together with an CNN feature extractor, the multiple labels for an image $\mb{x}$ will be predicted by
\begin{equation}\label{eq:predict}
	\mb{\hat{y}} = \sigma\left({\text{AGCN}(\mb{A}^t, \mb{H}^{t,0})}\otimes \text{CNN}\left(\mb{x}\right)\right),
\end{equation}
where $\mb{A}^t$ denotes the ACM and $\mb{H}^{t,0}$ is the initialized graph node. Prediction $\mb{\hat{y}} = [\mb{\hat{y}}_\text{old}~\mb{\hat{y}}_\text{new}]$, where $\mb{\hat{y}}_\text{old}\in\mbb{R}^{|\mc{C}_\text{seen}^{t-1}|}$ for old classes and $\mb{\hat{y}}_\text{new}\in\mbb{R}^{|\mc{C}^t|}$ for new classes. We train the current task for classifying using the Cross Entropy loss.

To mitigate the class-level catastrophic forgetting, inspired by the distillation-based lifelong learning method~\cite{li2017learning,zhou2022few}, we construct auto-updated expert networks consisting of CNN$_\text{xpt}$ and AGCN$_\text{xpt}$.
The expert parameters are fixed after each task has been trained and auto-update along with new task learning.
Based on the expert, we construct the distillation loss as
\begin{equation}\label{eq:distillation}
		\ell_\text{dst}({\mb{\hat{z}}},\mb{\hat{y}}_\text{old})=-\sum_{i=1}^{|\mc{C}_\text{seen}^{t-1}|}\left[{{\hat{z}_i}}\log\left({\hat{y}}_{i}\right)+\left(1-{{\hat{z}}}_{i}\right)\log\left(1-{\hat{y}}_{i}\right)\right],
\end{equation}
where ${\mb{\hat{z}}} = \sigma\left(\text{AGCN}_\text{xpt}(\mb{A}^{t-1}, \mb{H}^{t-1,0}) \otimes \text{CNN}_\text{xpt}\left(\mb{x}\right)\right)$ can be treated as the \textit{soft labels} to represent the prediction on old classes.
The $i$-th element $\hat{z}_i$ of ${\mb{\hat{z}}}$ represent the probability  that the image $\mb{x}$ contains the class.

To mitigate the relationship-level forgetting, we constantly preserve the established relationships in the sequential tasks. 
The graph node embedding is irrelevant to the label co-occurrence and can be stored as a teacher to avoid the forgetting of label relationships.
Suppose the learned embedding after task $t$ is stored as $\mb{G}^{t}=\text{AGCN}_\text{xpt}(\mb{A}^{t}, \mb{H}^{t,0})$, $t>1$.
We propose a relationship-preserving loss as a constraint to the class relationships:
\begin{equation}\label{eq:graph_distillation}
		\ell_\text{gph}({\mb{G}}^{t-1},\mb{H}^{t})=\sum^{|\mc{C}_\text{seen}^{t-1}|}_{i=1} \left\Vert{\mb{G}}^{t-1}_i-\mb{H}^{t}_i\right\Vert^2.
\end{equation}
By minimizing $\ell_\text{gph}$ with the partial constraint of old node embedding, the changes of $\text{AGCN}$ parameters are limited.
Thus, the forgetting of the established label relationships are alleviated with the progress of LML classification. 
The final loss for the model training is defined as
\begin{equation}\label{eq:final_loss}
	\ell=\lambda_1\ell_\text{cls}(\mb{y},\mb{\hat{y}}_\text{new})+\lambda_2\ell_\text{dst}(\mb{\hat{z}},\mb{\hat{y}}_\text{old})+ \lambda_3\ell_\text{gph}({\mb{G}}^{t-1},\mb{H}^{t}),
\end{equation}
where $\ell_\text{cls}$ is the classification loss, $\ell_\text{dst}$ is used to mitigate the class-level forgetting and $\ell_\text{gph}$ is used to reduce the  relationship-level forgetting. $\lambda_1$, $\lambda_2$ and $\lambda_3$ are the loss weights for $\ell_\text{cls}$, $\ell_\text{dst}$ and $\ell_\text{gph}$. Extensive  ablation studies are conducted  for $\ell_\text{gph}$  after all relationships are built. 


%% file: ex.tex
\section{Experiments}
\begin{figure}[t]
	\centering
	\includegraphics[width=\linewidth]{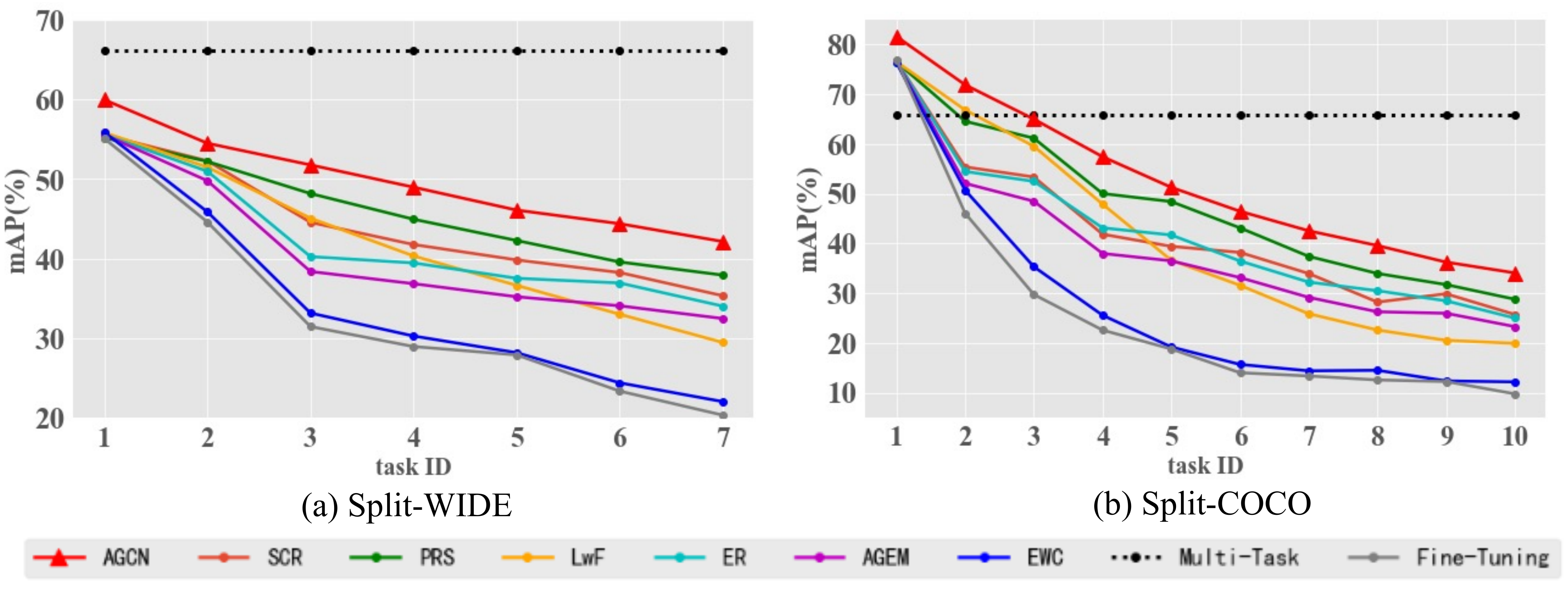}
	\vspace{-20px}
	\caption{mAP (\%) changes on two benchmarks.}
	\label{fig:learning_curves}
	\vspace{-15px}
\end{figure}

\begin{table}[t]
	\centering
	\caption{We report 3 main metrics (\%) for LML after the whole data stream is seen once on Split-WIDE and Split-COCO.}
	\resizebox{\linewidth}{!}{
	\begin{tabular}{c|rrr|rrr}
		\toprule
		\multirow{2}{*}{\textbf{Method}} & \multicolumn{3}{c|}{\textbf{Split-WIDE}} & \multicolumn{3}{c}{\textbf{Split-COCO}}  \\ 
		\cline{2-7} 
        \cline{2-7} 
		&\textbf{mAP} $\uparrow$&\textbf{CF1} $\uparrow$&\textbf{OF1} $\uparrow$&\textbf{mAP} $\uparrow$&\textbf{CF1} $\uparrow$&\textbf{OF1} $\uparrow$\\
		\hline
		\hline
		\textbf{Multi-Task}&66.17&61.45&71.57
		&65.85&61.79&66.27
		 \\

		\hline
		\textbf{Fine-Tuning}&20.33&19.10&35.72 
		&9.83&10.54&28.83
		\\
		\rowcolor{mygray}
		Forgetting $\downarrow$ &40.85&31.20&15.10
		  &58.04&63.54&20.60
		\\
		 \hline

		\textbf{EWC}&22.03&22.78&35.70
		&12.20&12.50&29.67
		
 \\
		\rowcolor{mygray}
		Forgetting $\downarrow$&34.86&28.18&15.17
		&45.61&55.44&19.85
		
 \\
		 \hline
		\textbf{LwF}&29.46&{29.64}&{42.69}
		&{19.95}&{21.69}&{40.68}
		 \\
		\rowcolor{mygray}
		Forgetting $\downarrow$&20.26&18.99&5.73
		&41.16&39.85&11.43
		 \\
		 \hline
		\textbf{AGEM}&32.47&33.28&38.93
		&23.31&27.25&37.94
		\\
		\rowcolor{mygray}
		Forgetting $\downarrow$&16.42&15.71&9.73
		&34.52&18.92&12.94
		 \\
		 \hline
		\textbf{ER}&{34.03}&{34.94}&{39.37}
		&25.03&30.54&38.38
		 \\
		\rowcolor{mygray}
		Forgetting $\downarrow$&15.15&11.80&8.61
		&33.46&17.28&12.34 
	 \\
		 \hline
		 \textbf{PRS}&{37.93}&{21.12}&{15.64}
		&{28.81}&18.40&13.86 \\
		\rowcolor{mygray}
		Forgetting $\downarrow$&13.59&51.09&62.90
		&30.90&54.36&52.51 \\
		\hline
		\textbf{SCR}&{35.34}&{35.47}&{41.92}
		&{25.75}&30.63&39.10 \\
		\rowcolor{mygray}
		Forgetting $\downarrow$&14.26&10.17&8.04
		&32.02&15.98&11.96 \\

		\hline
		\textbf{AGCN (Ours)}&\textbf{41.12}&\textbf{38.27}&\textbf{43.27}
		&\textbf{34.11}&\textbf{35.49}&\textbf{42.37}\\
		\rowcolor{mygray}
		Forgetting $\downarrow$&11.22&5.43&4.28
		 &23.71&14.79&8.16\\
		\bottomrule
	\end{tabular}}
\label{tab:COCO}
\vspace{-15px}
\end{table}

\begin{table}[t] 
    \centering
    \caption{Ablation studies (\%) for  ACM $\mb{A}^t$  used to model intra- and inter-task label relationships on Split-COCO.}
    \resizebox{0.8\linewidth}{!}{
        \begin{tabular}{c|cc|ccc}
        \toprule
            
             &$\mb{A}^{t-1}$ \& $\mb{B}^t$  & $\mb{R}^t$ \& $\mb{Q}^t$  &\textbf{mAP} $\uparrow$&\textbf{CF1} $\uparrow$&\textbf{OF1} $\uparrow$  \\
            \hline
           1& $\surd$ & $\times$ &  31.52 & 30.37 & 34.87   \\
           2& $\surd$ & $\surd$ & 34.11 & 35.49 & 42.37  \\
            \bottomrule
    \end{tabular}}
    \label{tab:ablation}
    \vspace{-15px}
\end{table}

\subsection{Dataset construction}

\noindent
\textbf{Split-COCO}.
We choose the 40 most frequent concepts from 80 classes of MS-COCO~\cite{lin2014microsoft} to construct Split-COCO, which has 65082 examples for training and 27,173 examples for validation.
The 40 classes are split into 10 different tasks, and each task contains 4 classes.

\noindent
\textbf{Split-WIDE}. 
NUS-WIDE~\cite{chua2009nus} has a larger scale than MS-COCO.
Following~\cite{jiang2017deep}, we choose the 21 most frequent concepts from 81 classes of NUS-WIDE to construct the Split-WIDE, which has 144,858 examples for training and 41,146 examples for validation.
We split the Split-WIDE into 7 tasks, where each task contains 3 classes. 




\subsection{Evaluation metrics}

\noindent
\textbf{Multi-label evaluation}. Following the traditional multi-label learning \cite{chen2019multi,kim2020imbalanced,chen2021learning}, 3 more important multi-label metrics: \textbf{mAP}, \textbf{CF1} and \textbf{OF1}.
\noindent
\textbf{Forgetting measure}~\cite{chaudhry2018riemannian}. 
This score denotes the value difference of the above three multi-label metrics between the final score and the score when the task was first trained done.

            

\begin{table}[t] 
    \centering
    \caption{AGCN ablation studies (\%) for loss weights and relationship-preserving loss on Split-COCO.}
    \resizebox{0.8\linewidth}{!}{
        \begin{tabular}{ccc|ccc}
            \toprule
             $\lambda_1$ & $\lambda_2$   &$\lambda_3$   &\textbf{mAP} $\uparrow$&\textbf{CF1} $\uparrow$&\textbf{OF1} $\uparrow$ \\
            \hline
            \hline
             
             $0.05$ & $0.95$   & $0$    & 29.90 & 31.80 & 37.12  \\
             \rowcolor{mygray}
              & Forgetting $\downarrow$&  & 29.24 & 24.88 & 19.67  \\
              \hline
             $0.07$ & $0.93$   & $0$    & 30.99 & 32.03 & 39.31   \\
             \rowcolor{mygray}
              &  Forgetting  $\downarrow$&    & 28.28 & 22.55 & 13.88   \\
              \hline

             $0.09$ & $0.91$   & $0$    & 29.71 & 32.71 & 38.91 \\
             \rowcolor{mygray}
              &  Forgetting  $\downarrow$&     & 29.97 & 21.79 & 16.49 \\
             \bottomrule
             $0.07$ & $0.93$   & $ 10^4 $    & 33.05 & 33.31 & 41.04   \\
             \rowcolor{mygray}
             &  Forgetting  $\downarrow$&     & 26.41 & 20.99 & 11.38 \\
             \hline
             $0.07$ & $0.93$   & $ 10^5 $    & \textbf{34.11} & \textbf{35.49} & 42.37   \\
             \rowcolor{mygray}
             &  Forgetting  $\downarrow$&     & 23.71 & 14.79 & 8.16 \\
             \hline
             $0.07$ & $0.93$   & $ 10^6 $    & 33.71 & 33.05 & \textbf{42.62}   \\
             \rowcolor{mygray}
             &  Forgetting  $\downarrow$&     & 25.69 & 21.30 & 7.89 \\
            \bottomrule
    \end{tabular}}
    \label{tab:lambda}
\vspace{-15px}
\end{table}

\begin{figure}[t]
	\centering
	\includegraphics[width=\linewidth]{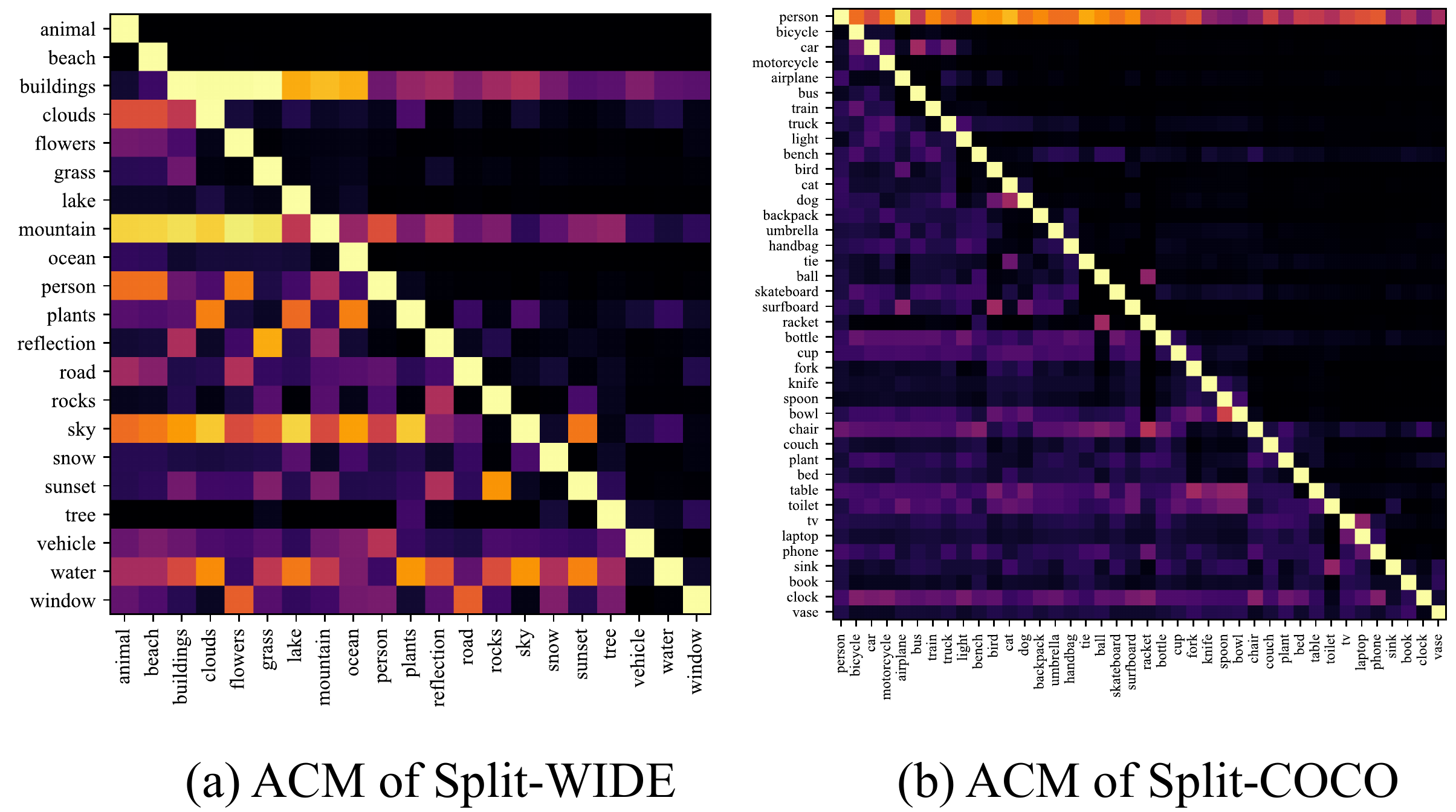}
        \vspace{-20px}
	\caption{ACM visualization on Split-WIDE and Split-COCO.}
	\label{fig:visualization_NUS}
	\vspace{-18px}
\end{figure}

\subsection{Results}


\textbf{Multi-Task} is the performance upper bound, and \textbf{Fine-Tuning} is the performance lower bound. 
In Tab.~\ref{tab:COCO}, our method shows better performance than the other state-of-art methods on the three metrics, as well as the forgetting value evaluated after task $T$. 
On Split-COCO, the AGCN outperforms the best of the state-of-art method PRS by a large margin (34.11\% vs. 28.81\%). 
The AGCN shows better performance than the others on Split-WIDE (41.12\% vs. 37.93\%), suggesting that AGCN is effective on the large-scale multi-label dataset.
As shown in Fig.~\ref{fig:learning_curves}, which illustrates the mAP changes as tasks are being learned on two benchmarks. 
The proposed AGCN is better than other state-of-art methods through the whole LML process.

The final ACM visualization is shown in Fig.~\ref{fig:visualization_NUS}.
The dependency between two classes with higher correlation has larger weights than irrelevant ones, which means the intra- and inter-task relationships can be well constructed even if the old classes are unavailable. 


\subsection{Ablation studies}
\textbf{ACM effectiveness}. 
In Tab.~\ref{tab:ablation}, if we do not build the relationships across old and new tasks, the performance of AGCN (Line 1) is already better than other non-AGCN methods, for example, 31.52\% vs. 28.81\% in mAP.
This means only intra-task label relationships are effective for LML. 
When the inter-task block matrices $\mb{R}^t$ and $\mb{Q}^t$ are available, AGCN with both intra- and inter-task relationships (Line 2) can perform even better in all three metrics, which means the inter-task relationships can further enhance the multi-label recognition. 

\noindent
\textbf{Hyperparameter selection}.
Then, we analyze the influences of loss weights and relationship-preserving loss on Split-COCO as shown in Tab.~\ref{tab:lambda}. 
When $\lambda_1=0.07$, $\lambda_2=0.93$, the performance is better than others. 
By adding the relationship-preserving loss $\ell_\text{gph}$, the performance obtains larger gains, which means the mitigation of catastrophic forgetting of relationships is quite crucial for LML classification. 
We select the best $\lambda_3$ as the hyper-parameters, \ie, $\lambda_3=10^5$ for LML classification.